\documentclass{article}
\usepackage{spconf,amsmath,epsfig}

\usepackage{tikz} 

\let\OLDthebibliography\thebibliography
\renewcommand\thebibliography[1]{
  \OLDthebibliography{#1}
  \setlength{\parskip}{0pt}
  \setlength{\itemsep}{0pt plus 0.3ex}
}

\usepackage{pdfpages} 

\begin{document}\sloppy

% Example definitions.
% --------------------
\def\x{{\mathbf x}}
\def\L{{\cal L}}

% Title.
% ------
\title{Learned Video Codec with Enriched Reconstruction \\for CLIC P-Frame Coding}
%
% Single address.
% ---------------
\name{David Alexandre$^1$, Hsueh-Ming Hang$^2$}
%Address and e-mail should NOT be added in the submission paper. They should be present only in the camera ready paper. 
\vspace{-2cm}
\address{Dept. of Electronics Engineering\\
National Chiao Tung University, Taiwan\\
$^1$\small{\underline{davidalexandre.eed05g@nctu.edu.tw}}, $^2$\small{\underline{hmhang@nctu.edu.tw}}
}

\vspace{-1.3cm}
\maketitle

\begin{abstract}
This paper proposes a learning-based video codec, specifically used for Challenge on Learned Image Compression (CLIC, CVPR Workshop) 2020 P-frame coding. More specifically, we designed a compressor network with Refine-Net for coding residual signals and motion vectors. Also, for motion estimation, we introduced a hierarchical, attention-based ME-Net. To verify our design, we conducted an extensive ablation study on our modules and different input formats. Our video codec demonstrates its performance by using the perfect reference frame at the decoder side specified by the CLIC P-frame Challenge. The experimental result shows that our proposed codec is very competitive with the Challenge top performers in terms of quality metrics.

\end{abstract}

\section{Introduction}
\label{sec:intro}

In the past three decades, video coding has been a popular research topic in multimedia signal processing. For the entertainment and communication industries, the need for better video coding scheme is very desirable to cut down the transmission and storage cost. Even though there are well established video coding standards such as H.265 and H.266 \cite{sullivan2012overview} \cite{VVC}, video coding research is always ready for its next generation. With the advancement of machine learning, the research on video coding entering a new era. The learning-based schemes integrate deep learning (DL) techniques to further enhance or even replace the existing components such as intra frame coding and inter frame coding. Lu, \emph{et al.}~\cite{lu2019dvc} initiates the end-to-end scheme for learning-based video coding. The main components inside the conventional video codec, such as motion estimation and residual coding, are replaced by the trained neural-network models and the learned codec produces results close to H.265 in their evaluation.

In the past 3 years, an annual workshop has been held in CVPR called Workshop and Challenge on Learning Image Compression (CLIC) \cite{clic2020}. Since 2018, CLIC hosted low-rate lossy image compression challenge sponsored by several well-known research institutes. In 2020, CLIC launched a new challenge item for P-frame coding. Given the reference frame and the target frame, the participant must compress video frames at around 0.075 bpp in average. On the leaderboard, the performance of submissions is compared using the MS-SSIM metric. This competition has proven to be able to gather submissions with interesting approaches, promoting the advancement of research in learning-based image/video coding. 

As one of the participates in CLIC2020, we expand our previous P-frame coding proposal significantly and present our refined video coding system in this paper. First, a compressor is designed to encode the sparse residual image signals and motion vector maps. Second, we introduce a refinement network (Refine-Net) to enhance the reconstructed image quality. It is an integral part of our compressor. Unlike the conventional residual signal compressor, which tries to approximate only the frame difference, our compressor generates some side information, which can assist the Refine-Net to reconstruct the target image. Third, we propose a cost hierarchical motion estimator with local-attention, which produces smooth (low coding rate) optical fields with low computational complexity. Forth, we enable a bypass coding mode, which disables motion compensation in the case that motion compensation does provide coding advantage. If an acronym is needed, our system may be called Deep Enriched Video Compressor, or DEVC in short. Compared to the top teams on the CLIC2020 leader board, our method shows competitive results and produces similar reconstruction quality.

In Section \ref{sec:related_work}, we review the related work in deep video coding. Section \ref{sec:proposed_method} describes our system framework, its components, and the training process. In Section \ref{sec:experiments}, we provide the dataset details, the evaluation setup, and the evaluation result, and we discuss and conclude our scheme in Section \ref{sec:conclusion}. Limited by space, additional information is included in the Appendix. 

\begin{figure*}
  \centering
  \includegraphics[width=0.75\textwidth]{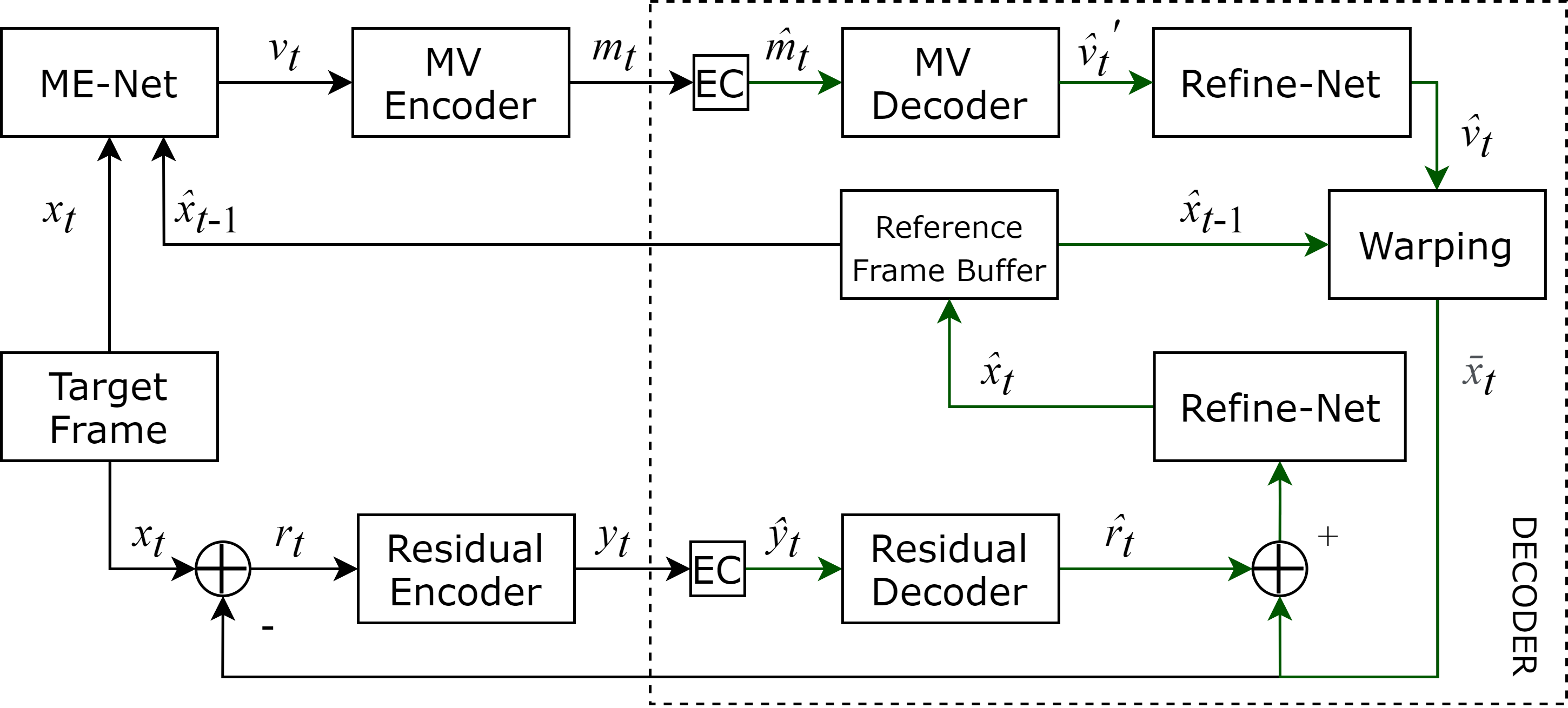}
  \caption{Our framework for P-frame Coding. It mainly consists of the motion estimation and compensation part and the residual coding part. Noted that the proposed Refine-Net is used together with Motion Compressor and Residual Compressor. The image luminance and chrominance components are separately processed by similar Motion-compensated Wrapper and Residual Compressor. But motion vectors are estimated using only the luminance component and the motion information is shared by the chrominance components. The decoder system is a subset of the encoder enclosed by the dashed lines. The green arrows inside the dashed lines indicate the decoding process at the decoder.}
  \label{fig:architecture-img}
  \vspace{-0.4cm}  
\end{figure*}

\vspace{-0.2cm}
\section{Related Work}
\label{sec:related_work}
\vspace{-0.2cm}
There are a number of video coding papers including DL technologies published recently. Some papers use DL techniques to optimize the selection of parameter values in a traditional video codec (such as H.266). Here, we pay attention only to the end-to-end Deep Neural Net (DNN) based video codec, in which the backbone system is mostly DNN.

DVC \cite{lu2019dvc} is one of the earlier DL-based video coding schemes. It adopts the structure of the traditional motion-compensated transform codec but replaces its components by the DL-based counterparts. For example, the conventional block motion estimator is replaced by a DL-based optical flow estimator (Flow Estimation Network). The conventional transform codec is replaced by a CNN-based compressor (Residual Coding Network). In the coding process, its Group of Picture (GOP) contains only I-frame and P-frames. The other recent DL-based video codecs often adopt a similar structure of DVC but have their own enhanced building blocks. For example, M-LVC \cite{lin2020m} suggests multiple-frame motion compesation and motion vector prediction. Agustsson, \emph{et al.}~\cite{agustsson2020scale} proposes a scale-space flow network to handle the challenging issues of disocclusion and fast motion in motion estimation/compensation. To reduce motion vector coding bit rate, Hu, \emph{et al.}~\cite{Hu2020ImprovingDV} designs a multi-scale resolution-adaptive flow coding scheme. Using a different approach, Ho, \emph{et al.}~\cite{Ho2020b} incorporates the classical parametric overlapped block motion compensation (POBMC) and reduces the motion vector number by a learning algorithm. Djeoulah, \emph{et al.}~\cite{djelouah2019neural} and Yang, \emph{et al.}~\cite{yang2020learning} introduce B-frame coding scheme in a learning-based video framework.

In most of the above published works, the residual or motion vector compressors (codec) adopt the still image codec proposed by Balle, \emph{et al.}~ \cite{balle2018variational} or Minnen, \emph{et al.}~\cite{minnen2018joint}. However, both the residual images (motion-compensated frame difference, MCFD) and the motion field (optical flow map, motion vector map) are information sparse signals. We believe a DL-based compressor specifically designed for coding residual signals would be more efficient than the existing solutions. Indeed, coupled with a Refine-Net, our compressor can generate “side information” to be used by Refine-Net at low bit rates. In addition, in this study, we improve the motion estimation and compensation networks. Different from the previous video codec of fixed coding path, we introduce a switchable bypass coding mode to improve individual coding efficiency.

\vspace{-0.4cm}

\section{Proposed Method}
\label{sec:proposed_method}
\vspace{-0.2cm}

\subsection{Overall Architecture}

\vspace{-0.2cm}
As shown in Fig. \ref{fig:architecture-img}, the our proposed system pipeline is similar to that of DVC \cite{lu2019dvc}. However, there are many differences. Our unique proposals are the residual and motion vector compressor network (Encoder and Decoder), the ME-Net, the Refine-Net, and the Bypass Coding Mode.

\begin{figure*}
  \centering
  {
  \includegraphics[width=1\textwidth]{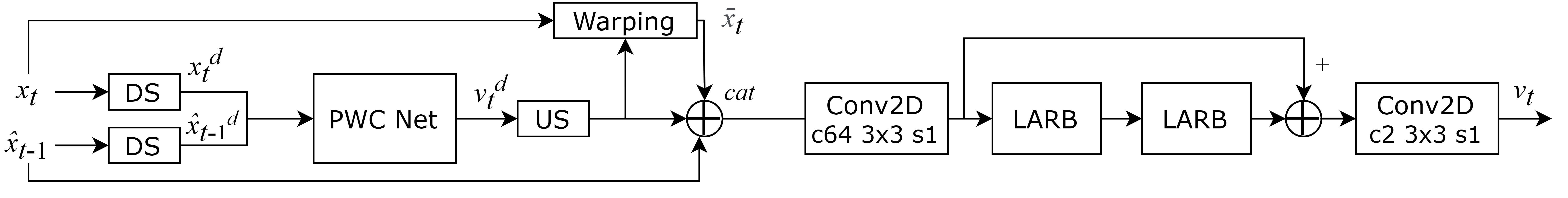}
  }
  \vspace{-0.6cm}
  \caption{Our ME-Net uses PWC-Net to estimate the motion vector $v_t$. We further enhance it by using local attention residual blocks (LARB). Note that a downsampled PWC-Net together with a hierarchical upsampling network is designed to improve motion field quality and to speed up the MV calculation.}
  \label{fig:me-net-img}
  \vspace{-0.5cm}
\end{figure*}

The entire encoder (Fig.\ref{fig:architecture-img}) has two main parts. The upper part is motion vector estimation (ME-Net), motion vector coding (MV codec), MV Refinement net (Refine-Net) and reference frame warping (Warping). The lower part is residual signal coding (Residual codec) and reconstructed frame refinement (Refine-Net). The part enclosed by the dashed lines is the decoder (receiver) system, which is a subset of the encoder. The EC box in the codec is Entropy Codec. In our design, the MV compressor network is applied to the motion vectors coming out of ME-Net. And the Residual Codec compresses the motion compensated frame differences (MCFD). However, in case of the Bypass Coding Mode, the motion vectors are not used, only the frame differences (FD) are coded and transmitted. 

It is worth to mention that the input color image is represented in the YUV (or YCbCr) format (the format used in CLIC20 P-frame Challenge). The 1D spatial resolution (image size) of U and V channels are half of that of Y. Each color channel is processed separately. The Y (luminance) channel is first processed using the entire system to produce the Y residual signals and the motion field, and the latter is to be used by all Y, U and V channels. The U and V channels are processed by two separate systems, which exclude the ME-Net and MV codec in Fig.\ref{fig:architecture-img}. Although Y, U, and V are processed independently (except for the shared motion field), but their neural nets share the same parameters (weights).

\vspace{-0.3cm}
\subsection{Motion Estimation}
Our framework uses a learning-based motion estimator, PWC-Net \cite{Sun2018PWC-Net}. To improve the extracted motion vector quality, we adopted the local attention residual block (LARB) proposed by Zhang, \emph{et al.}~\cite{zhang2019residual}. As shown in Fig. \ref{fig:me-net-img}, given the inputs of reference \(x_{t-1}\)  and target frame  \(x_t\), PWC-Net\((x_{t-1}, x_t)\) produces motion vector  \(v'_t\). Then, \(v'_t\) passes through the motion enhancement network (MV Refine-Net) to produce the refined motion vector  \(v_t\). In the training phase, we calculate the loss function using  \(MSE(Warp(x_{t-1}, v_t ), x_t)\). To speed up PWC-Net in computing MVs, the input images to PWC-Net are downsampled to (1/8) x (1/8) of its original size. To get back to the full size, we designed a learning-based multi-step   hierarchical upsampling process. In the training process, we average the distortion loss at all levels. The hierarchical network also offers a smoother motion field using fewer bits to code. Its detail is described in the Appendix.

The Warping operation is simply using the coded backward MVs, \(\hat{v}_t\), to construct an estimated frame, \(\bar{x}_t\). That is, every pixel of \(\bar{x}_t\) comes from \(\bar{x}_{t-1}\) guided by its associated coded MV.

\vspace{-0.2cm}
\subsection{Residual Coding}
\begin{figure}[ht!]
  \centering
  \includegraphics[width=1\columnwidth]{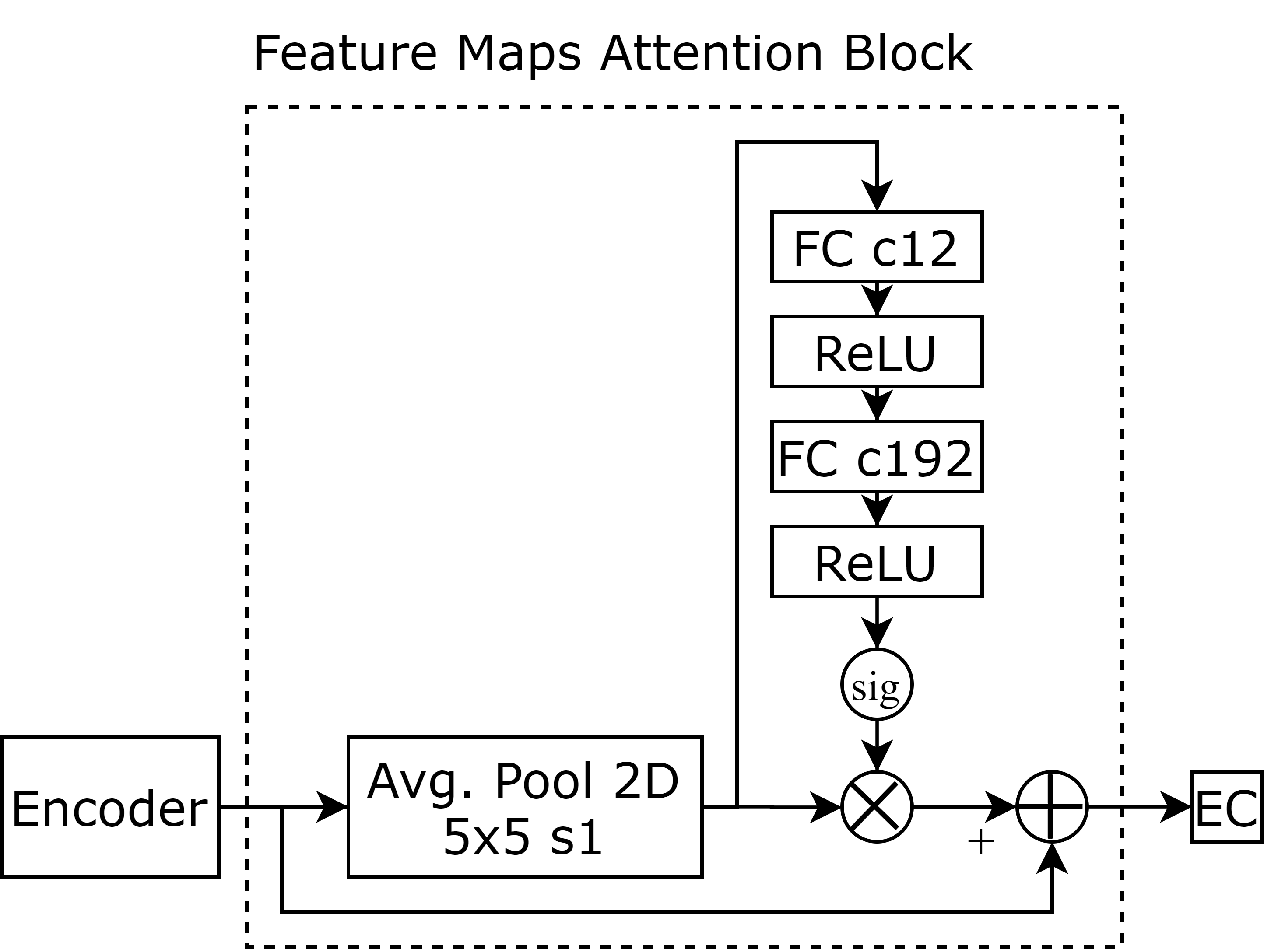}
  \caption{The attention block lies after the encoder and to optimize the feature maps only before the quantization process.}
  \label{fig:fmab-img}
  \vspace{-0.3cm}
\end{figure}

Residual images are sparse signals. The motion field is also information-sparse in the sense that the object and the background often have rather uniform values. Therefore, we design an CNN architecture aiming at compressing the sparse signals. Fig. 11 (Appendix) shows the architecture of the proposed system. The main backbone is similar to that of Balle, \emph{et al.}~\cite{balle2018variational}. However, we include an attention block, which is critical in picking up the sparse signals. With fewer internal channels (64) and the attention block, the compressor performs better than the original model with 128 internal channels. In addition, we found that for estimating the probability used by the Entropy Coder, the network proposed by Cheng, \emph{et al.}~\cite{cheng2020learned} is more effective. Then, the entire compressor is finetuned to cope with our target sparse signals. Two separate models are trained for coding residual image signals (MCFD) and motion field, separately.
The inputs to the residual compressor are in the single channel format mentioned earlier. The input values are normalized to (-1, 1). The encoder produces 128 output feature maps. An attention block (Fig. \ref{fig:fmab-img}) is placed at the end of the encoder to reduce the channel number to 64 (latent variables) and it also enhances the selection of sparse signals. Thus, the decoder has 64 channel input channels. We show the attention layer effect on the feature maps in Fig. 7 in the Appendix.
Similarly, the enhanced motion vectors are compressed using a compressor of the same structure as the residual codec. The MV encoder also has 64 output channels (similar to that of residual codec). However, the input to the MV compressor has only two channels because a MV has only the horizontal and the vertical displacements. The input value is normalized between 0 to 1. 

\vspace{-0.15cm}
  \vspace{-0.3cm}

\vspace{-0.3cm}
\subsection{Refine-Net}
\vspace{-0.15cm}
Our Refinement net (Refine-Net) is not a simple noise (artifacts) reduction post-processor as proposed by many other deep video codecs. When paired with our residual compressor, the compressor sends extra signaling to hint the Refine-Net to reconstruct a higher quality target frame. That is, by incorporating Refine-Net, the compressor learns to use the Refine-Net together to produce the target frame. Thus, our Refine-Net is an indispensable counterpart of the compressor. This specific function is achieved by a carefully designed 3-step training process. The compressor is first trained without the Refine-Net. Then, both Refine-Net and compressor are trained together to produce the desired property.

\begin{figure}[hb!]
  \centering
  \includegraphics[width=0.75\columnwidth]{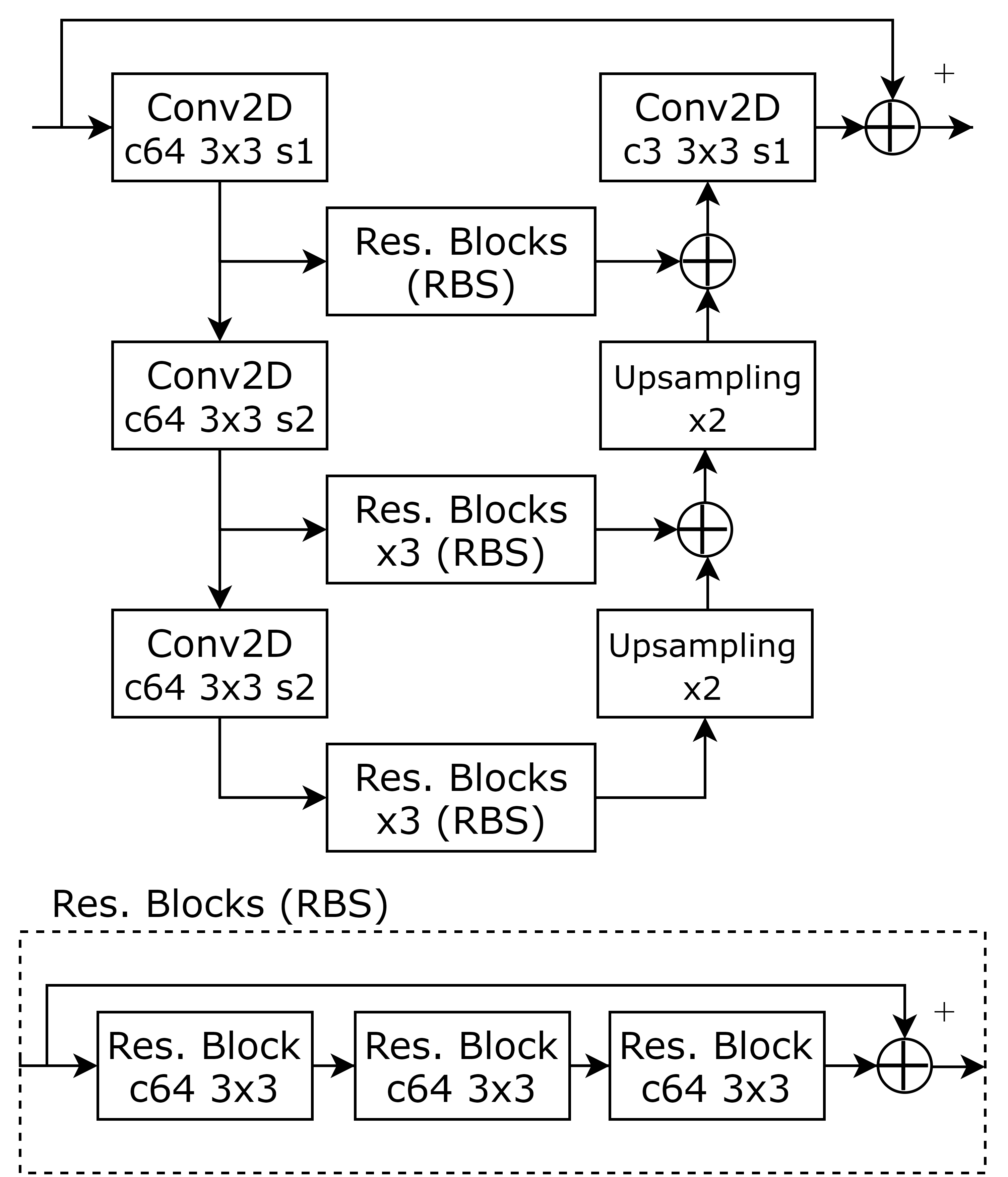}
  \caption{The Refine-Net has a multi-scale structure using the residual blocks. c64, 3x3, s2 means 64 channel and 3x3 convolution with stride 2.}
  \label{fig:refine-net-img}
  \vspace{-0.3cm}
\end{figure}

\begin{figure}[ht!]
\centering
    {
    \begin{tabular}{ccc}
    \(\hat{r}_t\) & \(\bar{x}_t + \hat{r}_t\) & \(\hat{y}_t\)\\

    \includegraphics[width=0.29 \columnwidth]{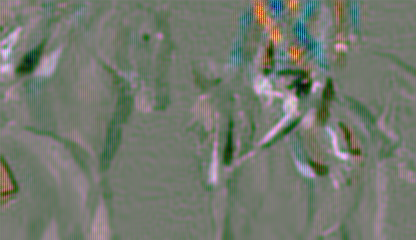}&
    \includegraphics[width=0.29 \columnwidth]{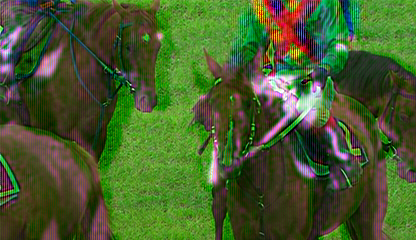}&
    \includegraphics[width=0.29 \columnwidth]{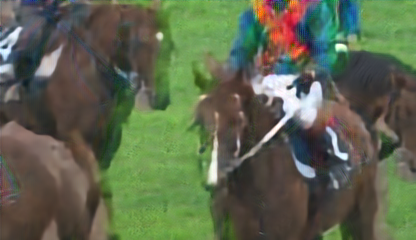}\\
    \\
    \includegraphics[width=0.29 \columnwidth]{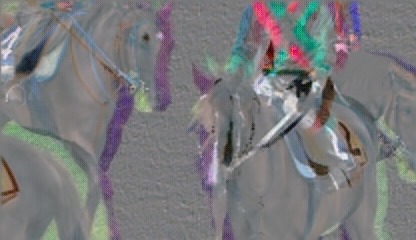}&
    \includegraphics[width=0.29 \columnwidth]{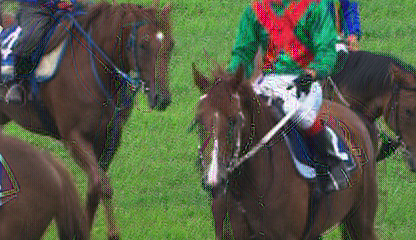}&
    \includegraphics[width=0.29 \columnwidth]{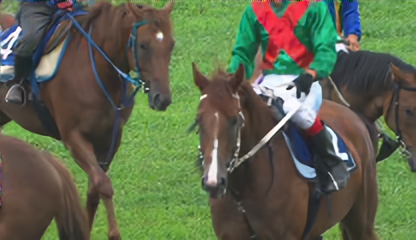} \\
    (a) & (b)& (c)\\
    \end{tabular}

    \caption{Upper Row - residual compressor with Refine-Net at low rates, and Lower Row - residual compressor with Refine-Net at higher rates. (a) The reconstruction of residual signal, (b) Intermediary result before Refine-Net, (c) Reconstructed frame after Refine-Net.}
    
     \label{fig:compare-refine-net-img}
    }
    %\vspace{-0.1cm}

    \vspace{-0.3cm}
\end{figure}

\begin{figure}[hb!]
\centering
{
\begin{tabular}{ccc}
\(v_t\) & \(\bar{v}_t\) & \(\hat{v}_t\) \\
\includegraphics[width=0.29 \columnwidth]{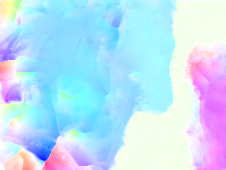}&
\includegraphics[width=0.29 \columnwidth]{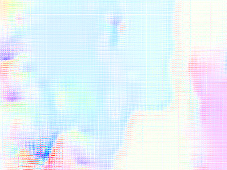}&
\includegraphics[width=0.29 \columnwidth]{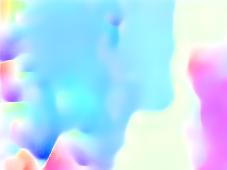}\\

(a) & (b)& (c)\\
\end{tabular}

\caption{Motion compressor with Refine-Net: (a) The original motion vector map, (b) Decoder output, (c) Reconstructed motion vector map (BPP: 0.05 bpp, PSNR: 27.86).}

\label{fig:compare-refine-net-motion-img}
}
%\vspace{-0.1cm}

\vspace{-0.3cm}
\end{figure}

We demonstrate the effect of Refine-Net in Figs. \ref{fig:compare-refine-net-img} and \ref{fig:compare-refine-net-motion-img}. At low bit rates (Fig. \ref{fig:compare-refine-net-img}, Upper Row) when few bits can be allocated to the residual signals, it is clearer to see that the residual image is not simply the MCFD. Some green-tone regions in Fig. 5(a) indicate the object that were in the reference frame \(x_{t-1}\) but not in the target frame, xt. Some red-tone regions indicate the opposite. Thus, the warped frame plus residual  \((\bar{x}_t + \hat{r}_t\) in Fig\ref{fig:compare-refine-net-img} \ref{fig:compare-refine-net-img}(b) is different from the target frame. However, the associated Refine-Net can take proper actions add/subtract objects to produce pretty good quality reconstructed image, as shown in Fig.\ref{fig:compare-refine-net-img}(c). In contrast, at high bit rates (Lower Row in Fig.\ref{fig:compare-refine-net-img}), the residual image is close to the original MCFD; hence, the warped frame plus residual image, Fig.\ref{fig:compare-refine-net-img}(b), is very close to the one after refinement, Fig. \ref{fig:compare-refine-net-img}(c). In other words, since the coded residual signal matches the original residual well, the Refine-Net does not alter much of its input.     
In the case of motion field (Fig. \ref{fig:compare-refine-net-motion-img}), the proposed Refine-Net can enhance the output quality by increasing motion vector accuracy. Fig. \ref{fig:refine-net-img} shows the design of Refine-Net. It contains a set of residual blocks. 

\vspace{-0.25cm}
\subsection{Quantization and Entropy Coding}
\vspace{-0.15cm}
We adopted the range coder of Duda, \emph{et al.}~\cite{duda2013asymmetric} (a version of arithmetic coder) as our lossless Entropy Codec. The feature maps of motion field or residual signal are first quantized using a binary quantizer for each feature map (channel). The quantizer threshold value and dequantizer reconstruction levels are based on the mean and variance produced by the context predictor in the hyperprior module defined by Eq. \ref{eq:decoder}, where \(\hat{y}^k\) is the reconstructed level at sample $k$, \(b^k\) is the transmitted binary bit stream drawing values from {-1,+1}, \(\sigma^k\) and \(\mu^k\) are the outputs of the context predictor (Fig. 11).
\vspace{-0.2cm}
\begin{equation}
   \begin{gathered}
       \sigma^k, \mu^k = {ctx\_predictor}(codes < k),\\
       \hat{y}^k = \mu^k + {\sigma^k}b^k\\
    \end{gathered}
    \label{eq:decoder}
    \vspace{-0.1cm}
\end{equation}

Then, the Entropy Encoder coverts the quantized signals to a binary stream. Because the quantization and entropy coding process is a non-differentiable operation, in the training process, we estimate the bit rate using the Gaussian-mixture model proposed by \cite{cheng2020learned} (Fig. 11 in Appendix) and performs entropy encoding and decoding only during the inference (testing) phase.

  \vspace{-0.3cm}

\subsection{Training Procedure}
\vspace{-0.15cm}
We train the motion estimator and the compressors separately. To train the motion estimator, we define the loss $L_{ME}$ as the average MSE of the warped frame at all levels as defined by Eq.(\ref{eq:me-net}), $d$ is the number of levels in Fig. 10. To be noted, the MV compressor using distortion loss \(D = MSE(x,\hat{x}\)).
\vspace{-0.2cm}
\begin{equation}
   L_{ME} = \frac{\sum\limits_{i=1}^d{MSE(warp(x_{t-1}^i, v_t^i), x_t^i)}}{d}
    \label{eq:me-net}
    \vspace{0.1cm}
\end{equation}

The loss function in training the compressor is the usual rate-distortion loss, \(L = \lambda * D + R\). As said earlier, we train our Compressor/Refine-Net models in three steps. In the first step, we exclude the Refine-net and only train the compressor networks to reconstruct the original inputs. The distortion term is calculated using \(D = MSE(x, \hat{x})\). The rate term $R$ is calculated from the Gaussian-mixture hyper prior model. In the second step, we freeze the compressor and train the Refine-Net using MSE. In the last step, we train the Compressor and the Refine-Net together in the end-to-end manner by changing the distortion term for \(D = 1-MSSSIM(x, \hat{x})\). The model is trained with high $\lambda$ in the early training. We switch to the target $\lambda$ after the model reaches convergence.

\vspace{-0.4cm}

\section{Experiments}
\label{sec:experiments}

\vspace{-0.2cm}
We evaluate our system performance using CLIC 2020 P-frame compression scenario. Given a perfect reference image frame and a target image frame, we use the proposed scheme to code one P-frame. In case where the motion field fails to do motion compensation well (i.e., the squared MCFD is larger than the squared FD for that frame), we send a Bypass Mode code (from the encoder) to the decoder to do the frame difference decoding and reconstruction without using the motion compensated frame. There are roughly 15-20\% frames using the Bypass Mode, mostly, animation, graphics, and games.  
\vspace{-0.2cm}

\vspace{-0.3cm}
\subsection{Experimental Setting}
\subsubsection{Training Setup}
We used Vimeo-90k \cite{xue2019video} triplet dataset to train our model. It contains 73,171 3-frame sequences with a fixed resolution of 448 x 256. During the training, each RGB image is cropped and flipped randomly into 256x256 and is converted to YUV format. To avoid over-fitting, we did not use UVG dataset in training, because its subset is used as the test dataset specified by CLIC. We reserve more bits for sending residual information; therefore, $\lambda$ = 128 in training the motion compressor in MSE. On the other hand, to meet the bytes limit, we empirically adjust $\lambda$ in the final training step. We set $\lambda$ = 1024 for training the residual compressor with MSE loss at first step and use $\lambda$ = 128 for MS-SSIM loss before it is adjusted to achieve the target bpp. 

\vspace{-0.3cm}
\subsubsection{Evaluation Method}
In the evaluation, we use CLIC 2020 P-frame Test Set. The dataset has 12,786 selected video frames drawn from the UGC Dataset without HDR video, vertical video, interlaced video, and smaller than 720p videos. Each frame is extracted using \emph{ffmpeg} tool and converted to three separated channels, Y, U, and V following the YUV420 format. That is, the U and V image resolution (1D) is half of that of Y.

CLIC selects MS-SSIM as the quality measure of the reconstructed frames. To ensure fair evaluation, we follow the method used in the CLIC as given in Eq.\eqref{eq:ms_ssim_eq}, where $size_i$ is the image size (pixel number) of the $i$-th frame. That is, given the total number of frames, $n$, in the dataset, we multiply the MS-SSIM score of a particular frame $i$ with its image size, $size_i$, and then we divide the total MS-SSIM by the total frame sizes in the dataset. We do the same operation for the PSNR calculation. Our model targets at an average bit rate of 0.075 bit per pixel (bpp) or the total compressed files not exceeding 3,900,000,000 bytes.
\vspace{-0.2cm}
\begin{equation}
   MS-SSIM = \frac{\sum\limits_{i=1}^n{MS-SSIM_i * size_i}}{\sum\limits_{i=1}^n{size}_i}
    \label{eq:ms_ssim_eq}
\end{equation}

\vspace{-0.65cm}

\subsection{Experimental Results}
The performance for our scheme is shown in Table 1. We compare our MS-SSIM results with the top 5 performers on the CLIC leader board. Our result at 0.074 bits per pixel has MS-SSIM=0.99675 and PSNR=37.46. Compared to the top performers in CLIC, our proposed scheme competes rather well. The final training step in this model was trained using 2x TESLA-V100 in DDP (Distributed Data Parallel) mode with batch size 8 for 1,000 epochs.

\begin{table}[ht!]
\centering
\begin{tabular}{ |p{2.75cm}|c|c|c| } 
 \hline
  & Bytes & MSSSIM & PSNR\\
   \hline
 Ours & 38799603 & 0.99675 & 37.460 \\ 
 TUCODEC\_SSIM & 37871408 & 0.99681 & 37.360 \\ 
 TUCODECVIDEO & 38152157 & 0.99681 & 37.355 \\
 DAMO\_XG & 38720613 & 0.99676 & 41.547 \\
 IMCL\_MSSSIM & 37962951 & 0.99670 & 37.663 \\
 SR\_VCOR & 38831925 & 0.99659 & 40.914 \\
 \hline
\end{tabular}
\caption{Comparison with the CLIC 2020 P-frame top performers at MSSSIM and PSNR.}

\label{table:comparison-clic}
\end{table}

\textbf{Motion Vector and Residual Information Proportion}. Under our setting, the MV part is compressed in the range of 0.01 to 0.05 bpp and has acceptable quality to be used for motion compensation. Meanwhile, our residual information has a wider range and use more rate. Overall, our MV information uses 18\% of the total bits. 

\vspace{-0.5cm}
\subsection{Ablation Study}
We studied diffident combinations and settings of the components in our system.

\textbf{Refine-Net}. We validate the usefulness of Refine-Net by conducting an experiment with and without Refine-Net. Fig. 9 (in Appendix) shows that Refine-Net provides about 1dB PSNR improvement, even higher at lower bit rates in our chosen subset of HEVC Class D Test set.

\textbf{Input format}. The input format to the compressor network has an impact on the compression efficiency. Our experiments show that compressing Y, U, and V separately is more efficient than using three channels together as RGB or YUV as the input. The evaluation in Fig. 8 (Appendix) shows that the separate input-channel residual compressor, which is trained for Y+U+V, outperforms the model trained for RGB format and YUV format.

\textbf{Model Size}. We compare the model size of our compressor with the original architecture and channel setting from Balle, \emph{et al.}~\cite{balle2018variational}. With the additional Refine-Net, our model size is still smaller by about 1M. Table 2 in the Appendix shows a size comparison for the model. 

\textbf{Compressor Performance}. We tested our compressor with and without the attention-layer and the hyperprior from Cheng, \emph{et al.}~\cite{cheng2020learned}. The compressor architecture is shown in Fig. 11 (Appendix). The evaluation result is shown in Fig. 9 (Appendix). Compare to its predecessor, our compressor works particularly well at lower rates with fewer number of channels.

\vspace{-0.5cm}
\section{Conclusion}
\label{sec:conclusion}

\vspace{-0.2cm}

This paper proposed a learning-based video codec for P-frame coding, which includes our designs of compressor, refinement net, and hierarchical motion estimator. Particularly, we design a compressor network and Refine-Net for coding motion field and residual information, and we propose a hierarchical ME-Net with local attention. Because motion-compensated frame difference and motion field are sparse signals, our dedicated compressors show better compression efficiency than using the conventional still-image compressors. Our final result is on-a-par with the winner of the CLIC 2020 P-frame competition. 

\vspace{-0.3cm}

\vspace{-0.2cm}
\section{acknowledgement}
\vspace{-0.2cm}
This work is partially supported by the Ministry of Science and Technology, Taiwan under Grant MOST 109-2634-F-009-020 through Pervasive AI Research (PAIR) Labs, National Chiao Tung University, Taiwan. We would like to thank Prof. Wen-Hsiao Peng, National Chiao Tung University, for his guidance and valuable comments on this work.

\vspace{-0.3cm}

\bibliographystyle{IEEEbib}
\bibliography{icme2021template}

\newpage


\section*{Supplementary material}

# APPENDIX

**1. Attention Layer Effect on Encoder.** Figure 7 shows the effect of the attention layer on the residual signals in coding. We can observe at the feature map is strengthen after adding in the attention layer, which helps the decompressor to reconstruct better with fewer number of channels in the encoder output feature maps.

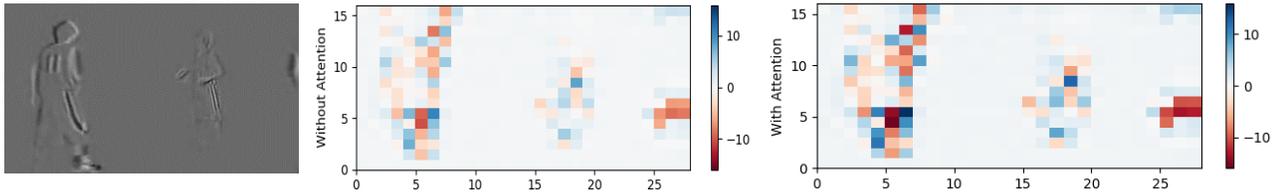

**Fig. 7.** The effect of attention layer on residual signals. The encoder output feature map without the attention layer (middle), the output feature map with the attention layer (right), and the decompressed residual (left), which comes the feature map with the attention layer (right).

**2. Input Format.** We compare different input formats for our model. The Y+U+V input model means each channel is compressed separately. The YUV (or RGB) input model means 3 channels are concatenated together to form the input to a single model. As Fig. 8 shows, the Y+U+V model has a higher MS-SSIM value.

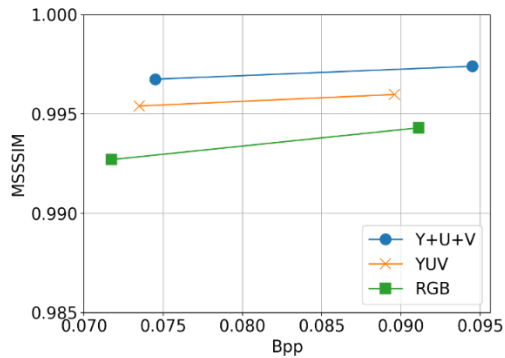

**Fig. 8.** Comparison of compressor input format. The PSNR evaluation on CLIC 2020 P-frame Test-Set. The Y+U+V model reaches the highest MS-SSIM compares to the YUV and RGB models.

**3. Model Size.** Our compressor model is compared with Balle, *et al.* [11] in model size. The number in Table 2 shows the parameter size of each compressor model.

|            | Ours    | Balle, *et al.*[11] |
|------------|---------|---------------------|
| Encoder    | 1.393M  | 3M                  |
| Decoder    | 2.702M  | 3M                  |
| Refine-net | 1M      | -                   |

**Table 2.** Comparison for our compressor model size. We save approximately 1M parameters (with Refine-Net).

**4. Compressor Performance.** Figure 9 compares our compressor performance with its predecessor (Balle, *et al.* [11]). With the help of attention layer and Refine-Net, ours performed better with fewer channel numbers.

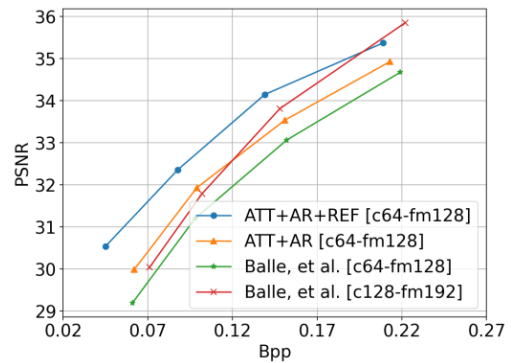

**Fig. 9.** Ablation study on the compressor architecture. It shows the PSNR and BPP of single P-Frame coding when using scale-hyperprior (SC) [11], attention block (AT) and auto-regressive hierarchical priors (AR) [15], attention block, auto-regressive, and refine-net (REF). c64 = 64 internal channels, fm128 = 128 output feature maps to quantizer. This test is conducted on a chosen subset of class D HEVC Test Set (RaceHorse, BasketBall, BlowingBubble).

**5. Hierarchical motion field generator (ME-Net).** Instead of using an optical flow net such as PWC Net to generate the full-size motion filed (motion vector map), we design a hierarchical motion field generator, PWC net plus a hierarchical upsampling structure. It offers the following advantages. (1) It produces a smoother motion field, which requires few bits to transmit (compared to the full-size flow net generated motion field). (2) With multiple-step enhancement, it often produces better quality motion field. (3) It reduces the computational complexity because a full-size flow net has very high complexity. Our motion field generator is shown in Fig. 10. The original current frame ($x_t$) and reference frame ($\hat{x}_{t-1} = x_{t-1}$) are down-sampled three times by the average pooling operator. The (1/8)x(1/8) frames, $x_t^{1/8}$ and $x_{t-1}^{1/8}$, are fed into the PWC Net to produce the motion field, $\bar{v}_t^{1/8}$. $\bar{v}_t^{1/8}$ is used to produce an intermediate warped image, $\bar{x}_t^{1/8}$. Then, $\bar{v}_t^{1/8}$ is enhanced by a simple Motion Enhancement Net to produce $v_t^{1/8}$. Note that the Motion Enhancement Net (1/8-level) takes in $x_t^{1/8}$ and the warped prediction, $\bar{x}_t^{1/8}$, in addition to $\bar{v}_t^{1/8}$. Hence, $v_t^{1/8}$ is a polished version of $\bar{v}_t^{1/8}$ to produce a better warped frame for $x_t^{1/8}$. Then, $v_t^{1/8}$ is upsample to $\bar{v}_t^{1/4}$, and then another Motion Enhancement Net (1/4-level) is employed to produce a polished motion field at 1/4-level, $v_t^{1/4}$. The same Motion Enhancement Net structure is repeated until we obtain the full-size motion field. In the training process, we use the motion field produced at each level to produce the warped frame and calculate the average MSE of that level. The average MSE of all levels is used to calculate the final loss function.

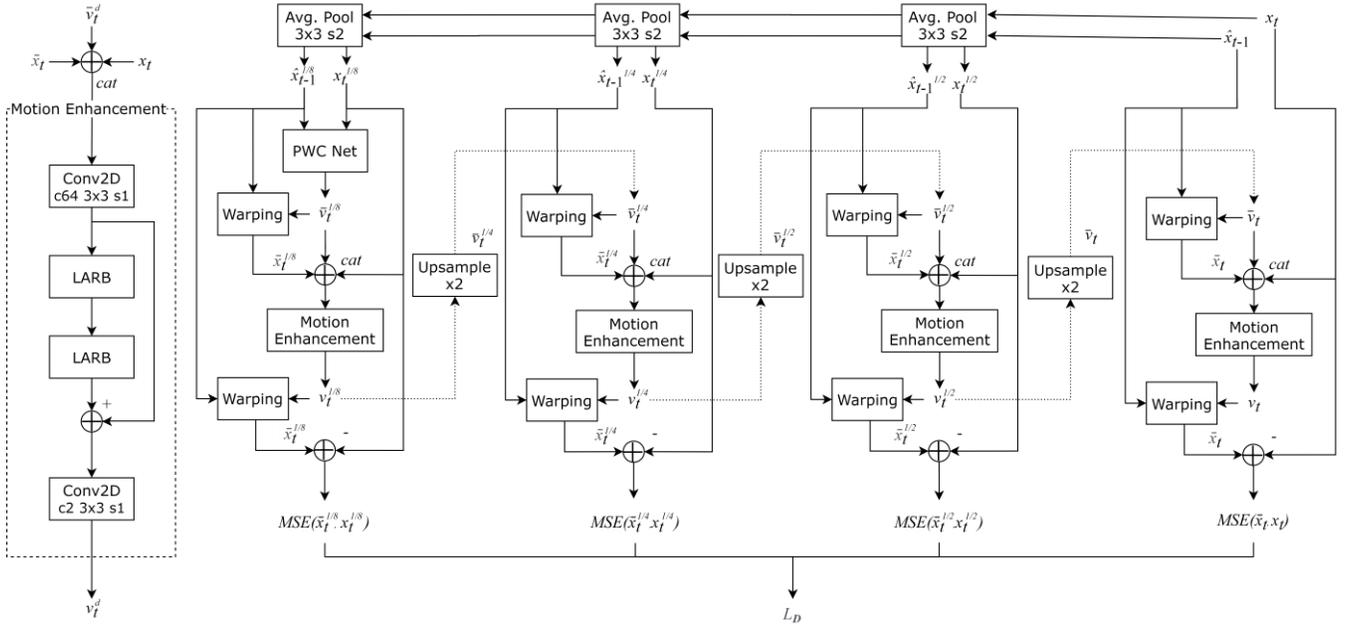

Fig. 10. Hierarchical motion field generator (ME-Net): The Motion Enhancement Net (left) is a module used in the hierarchical structure (right) of our ME-Net. The PWC-Net is performed on the 1/8 scale of the original frame size. To accommodate the loss information due to the 3-stage scaling, we construct a hierarchical enhancement structure to improve the motion field in the upscaling process. Our Motion Enhancement Net (left) uses two Local-Attention Residual Block. Note that the inputs to the Motion Enhancement Net is a concatenation of the warped frame (Y channel), the current frame (Y channel) and the estimated motion field. In training, the average MSE of every less are summed together to form the overall distortion.

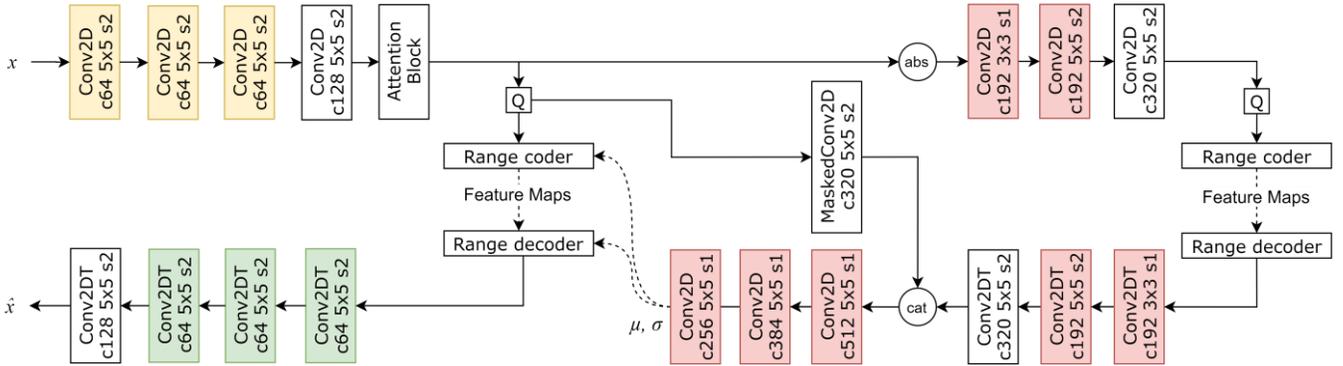

**Fig. 11.** Our compressor consists of basic codec (left half) and hyperprior module (right half). The basic encoder and decoder are similar to the codec of Balle, *et al.* [11]. We add an attention block into the end of the encoder. In training, our quantization is modeled as additive uniform noise. This architecture is used for coding the residual signals and the motion field. The hyperprior architecture is adopted from Cheng, et al. [15]. The yellow block has GDN activation. The green block has IGDN activation. The red block has ReLU activation.

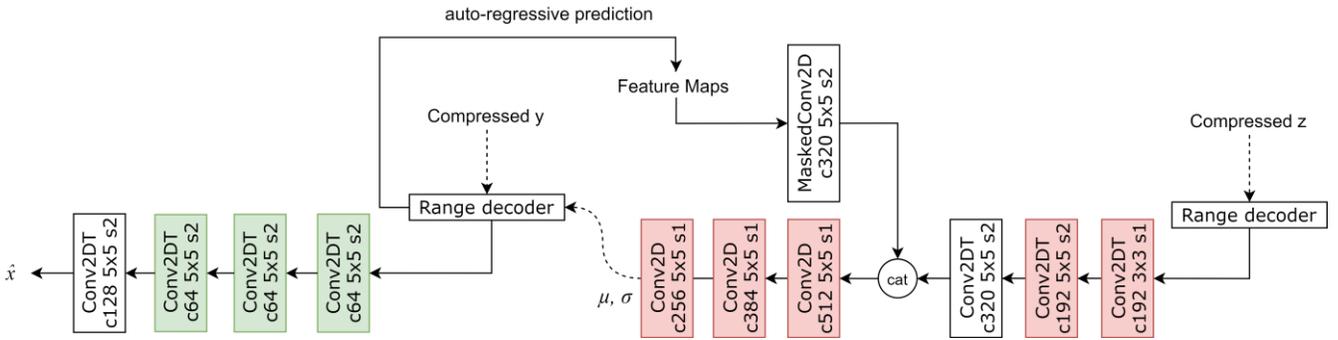

**Fig. 12.** The decoder part in our compressor. It is a subset of the complete codec (Fig.11). The hyperprior uses an auto-regressive mechanism to decode $y$ at the cost of computation.

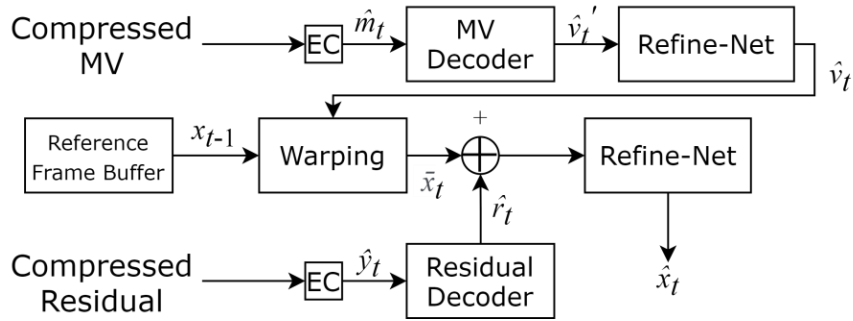

**Fig. 13.** The decoder for the CLIC P-Frame scenario. It needs two inputs, compressed motion field and compressed residual signals, to reconstruct the target $\hat{x}_t$. Noted that we use the perfect reference frame under the CLIC specification.

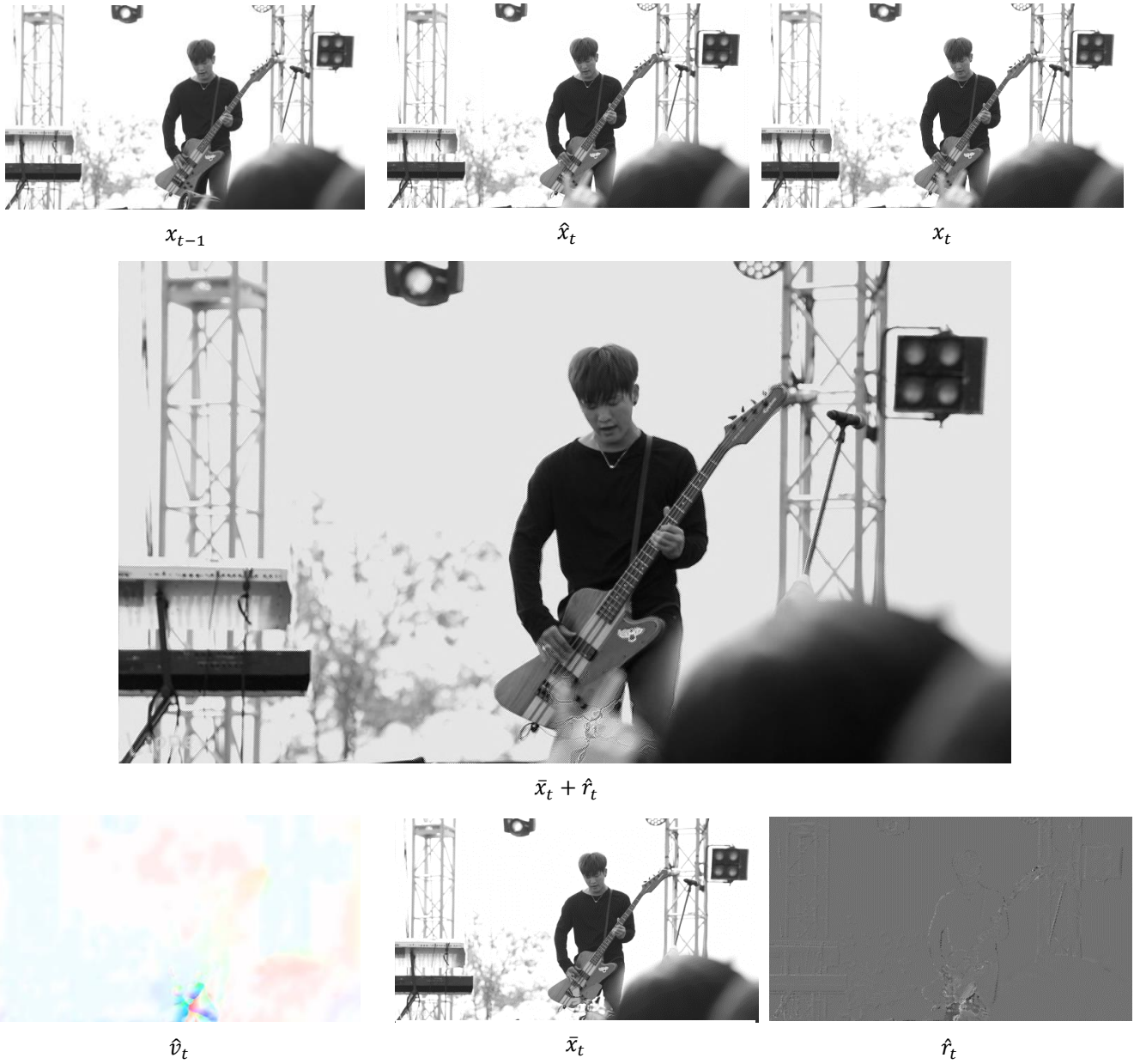

**Fig. 15.** Our decoding process uses the decompressed motion field $\hat{v}_t$, (0.01 bpp) to construct the motion compensated frame $\bar{x}_t$. We add the decompressed residual signals $\hat{r}_t$ (0.16bpp) and the motion compensated frame to produce the intermediary result $\bar{x}_t + \hat{r}_t$, and then it goes through the Refine-Net to generate $\hat{x}_t$ (PSNR = 44.26, MSSSSIM = 0.999). The reconstruction for U channel takes 0.065 bits-per-pixel (PSNR= 52.52, MSSSIM=0.999), and the reconstruction for V channel takes 0.056 bit-per-pixel (PSNR=51.81, MSSSIM=0.998). Only the Y-Channel images are shown here.

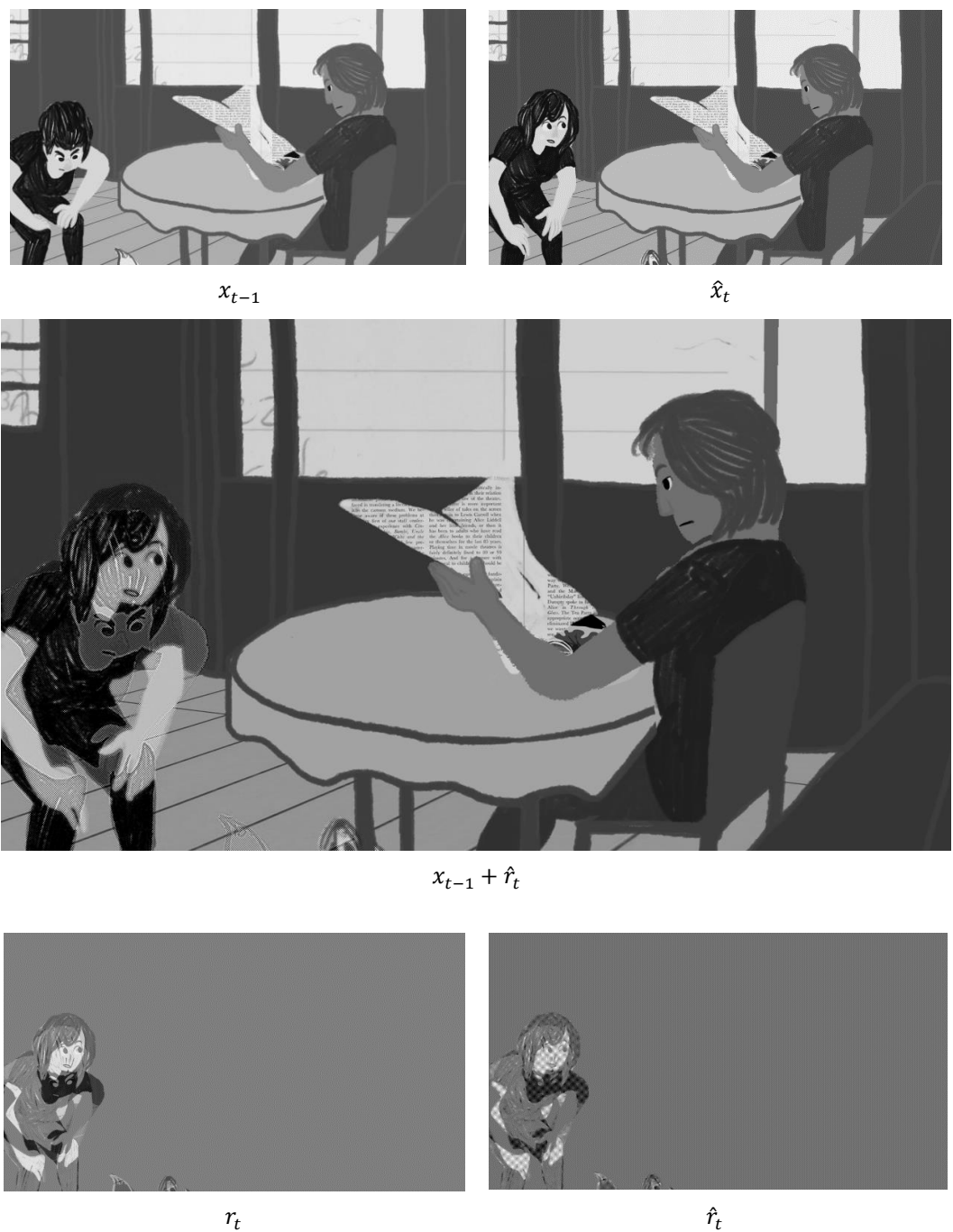

**Fig. 16.** Our decoding process skips the motion compensation when the motion estimation produces more error than without. It adds the decompressed residual signals $\hat{r}_t$ (0.16bpp) and the perfect reference frame $x_{t-1}$ to produce the intermediary result $x_{t-1} + \hat{r}_t$ and then it goes through the Refine-Net to generate $\hat{x}_t$ (PSNR = 43.18, MSSSSIM = 0.999). The reconstruction for U channel takes 0.034 bit-per-pixel (PSNR=54.92, MSSSIM=0.999), and the reconstruction for V channel takes 0.040 bit-per-pixel (PSNR=53.39, MSSSIM=0.999). Note that $\hat{r}_t$ contains extra signaling (grid pattern on face) to be used and removed by Refine-Net. Only the Y-Channel images are shown here.



\end{document}